\documentclass{article}

\usepackage[preprint]{neurips_2024}

\usepackage{natbib}
\setcitestyle{numbers,square}

\usepackage[utf8]{inputenc} 
\usepackage[T1]{fontenc}    
\usepackage{hyperref}       
\usepackage{url}            
\usepackage{booktabs}       
\usepackage{amsfonts}       
\usepackage{nicefrac}       
\usepackage{microtype}      
\usepackage{xcolor}         
\usepackage{wrapfig}

\usepackage{graphicx}
\usepackage{amsthm,amsmath,amssymb}
\usepackage{mathrsfs}

\title{Disentangling Foreground and Background Motion for Enhanced Realism in Human Video Generation}

\author{%
  Jinlin Liu\thanks{corresponding author}, Kai Yu, Mengyang Feng, Xiefan Guo, Miaomiao Cui\\
  \texttt{liujl09@live.com}\\
  \href{https://liujl09.github.io/humanvideo\_movingbackground/}{\textit{https://liujl09.github.io/humanvideo\_movingbackground/}}\\
}

\begin{document}
\maketitle

\begin{abstract}
    Recent advancements in human video synthesis have enabled the generation of high-quality videos through the application of stable diffusion models. However, existing methods predominantly concentrate on animating solely the human element (the foreground) guided by pose information, while leaving the background entirely static. Contrary to this, in authentic, high-quality videos, backgrounds often dynamically adjust in harmony with foreground movements, eschewing stagnancy. We introduce a technique that concurrently learns both foreground and background dynamics by segregating their movements using distinct motion representations. Human figures are animated leveraging pose-based motion, capturing intricate actions. Conversely, for backgrounds, we employ sparse tracking points to model motion, thereby reflecting the natural interaction between foreground activity and environmental changes. Training on real-world videos enhanced with this innovative motion depiction approach, our model generates videos exhibiting coherent movement in both foreground subjects and their surrounding contexts. To further extend video generation to longer sequences without accumulating errors, we adopt a clip-by-clip generation strategy, introducing global features at each step. To ensure seamless continuity across these segments, we ingeniously link the final frame of a produced clip with input noise to spawn the succeeding one, maintaining narrative flow. Throughout the sequential generation process, we infuse the feature representation of the initial reference image into the network, effectively curtailing any cumulative color inconsistencies that may otherwise arise. Empirical evaluations attest to the superiority of our method in producing videos that exhibit harmonious interplay between foreground actions and responsive background dynamics, surpassing prior methodologies in this regard.
\end{abstract}

\section{Introduction and Related Work}
The objective of image animation is to transform static images into dynamic videos, harnessing the original image's content. A prominent application lies in imbuing characters within input images with targeted actions, guided by reference footage \cite{siarohin2019first,siarohin2019animating,siarohin2021motion,yu2023bidirectionally,zhang2022exploring,zhao2022thin,wang2023disco,xu2023magicanimate,karras2023dreampose,hu2023animate}. Initial endeavors in this domain relied upon meticulously crafted loss functions alongside pure convolutional or transformer-based architectures to transpose the input image into varying poses \cite{siarohin2019first,siarohin2019animating,siarohin2021motion,yu2023bidirectionally,zhang2022exploring,zhao2022thin}. 

Recently, however, diffusion models have emerged as a generative strategy in works like Disco \cite{wang2023disco}, Magicanimate \cite{xu2023magicanimate}, Dreampose \cite{karras2023dreampose}, and Animateanyone \cite{hu2023animate}, capitalizing on pre-trained, high-performance models to yield outputs with heightened resolution and finer detail. Disco introduces ControlNet for background inpainting and dynamic foreground figure generation, while Dreampose integrates detailed cross-attention features from reference images and Densepose \cite{guler2018densepose} for refined guidance. Despite their advancements, these image-centric models encounter limitations in providing seamless frame-to-frame transitions for video generation. Magicanimate and Animateanyone tackle this issue by incorporating a temporal motion module \cite{guo2023animatediff}, utilizing additional U-Nets to capture intricate features for videos with meticulous detail.

Despite significant strides achieved in the past two years, we have identified an inherent shortcoming in image animation research that has hitherto received inadequate attention. The overwhelming majority of preceding efforts concentrate exclusively on animating the human subject within input images, transitioning it to various poses, while inadvertently overlooking the background. This oversight results in generated videos with consistently static background, a stark contrast to the reality observed in most online video content, where backgrounds are typically in motion or exhibit dynamism. Indeed, it is only in a minority of instances, typically those filmed with a stationary camera, that a truly static background is encountered. To enhance the authenticity of image-animated videos and bridge this gap, it becomes imperative to transcend the confines of static background generation. Incorporating dynamic backgrounds into synthesized videos is a pivotal step towards creating animations that more faithfully mimic the richness and variability of real-world footage, thereby elevating the overall realism and immersive quality of the output.

This research presents a groundbreaking method that isolates the modeling of foreground and background movement in video generation. We've developed an ingenious motion depiction system that skillfully captures intricate human actions along with environmental changes. Human activity is modeled using pose data, while background motion is traced via sparse tracking points. This separation allows our model to learn and create harmonious foreground-background interactions after being trained on real-world videos. To synthesize longer sequences without typical long-term errors, we employ a segmented generation technique. Videos are produced in sections, with each new clip informed by the final frame of the preceding one and fresh noise. To maintain continuity, the base reference image's features are continuously infused into the network throughout clip generation, preventing color inconsistencies from accruing as the video grows. In summary, our research advances the field in two pivotal areas:

- Our work pioneers a breakthrough in addressing and transcending the long-standing limitation of static backgrounds prevalent in image animation. By introducing a groundbreaking decoupling of foreground and background motion representation, we present the inaugural method capable of synthesizing human videos that not only feature naturalistic foreground actions but also incorporate authentically dynamic backgrounds, thereby setting a new benchmark for realism in generated video content.

- We introduce an efficient pipeline designed to produce extended videos devoid of error accumulation issues typically encountered in prolonged sequences. This is achieved through a strategic combination of conditional concatenation and the utilization of global feature extraction, enabling the seamless generation of prolonged video clips with maintained content consistency, thereby ensuring a cohesive and high-quality viewing experience throughout.

\section{Methodology}

\begin{figure}[h]
  \centering
  \resizebox{1\linewidth}{!}{
    \includegraphics{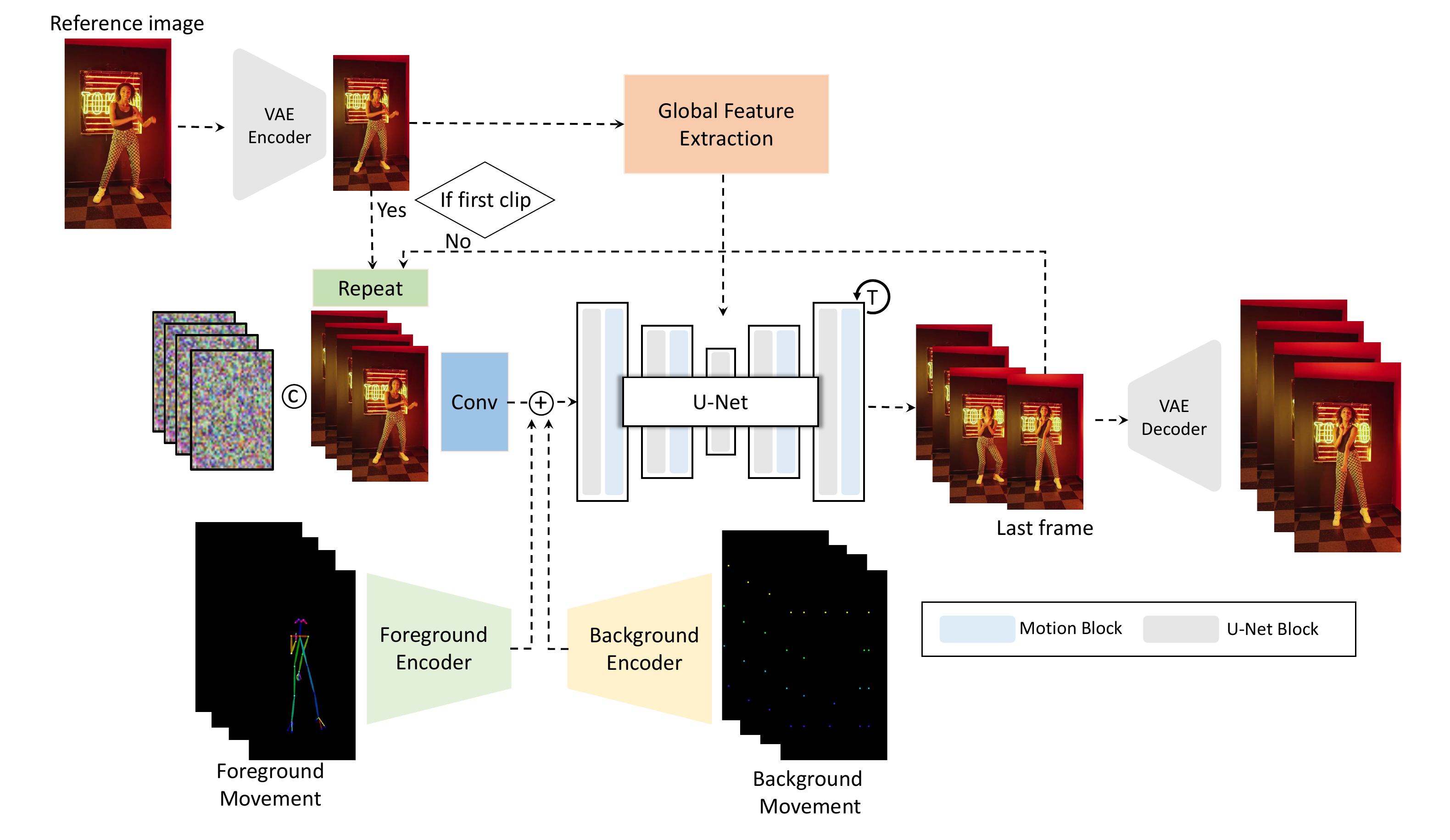} }
  \caption{ Our proposed method features a pipeline that meticulously models foreground motions with pose guidance and captures background dynamics via a sparse tracking system. To tackle the challenge of video synthesis beyond training sample lengths, we use a sequential strategy, generating video clips that build on the last frame of the previous clip to maintain continuity. A global feature vector derived from the initial image ensures visual coherence across the iterative process, preventing color or context inconsistencies. Additionally, a Temporal Motion Block ensures smooth transitions and temporal consistency between frames, enhancing realism and fluidity in the generated sequences. }
  \label{fig:pipeline}
\end{figure}

The methodology's workflow is illustrated in Figure \ref{fig:pipeline}, outlining our approach. We adopt a dual strategy: foreground movement is modeled leveraging pose direction, while background motion is handled through sparse tracking markers. To synthesize videos exceeding the length of our training frame number, we adopt a clip-by-clip generation technique. Here, the final frame of each produced clip serves as the springboard for generating the subsequent one. To uphold coherence throughout this sequential generation, the features extracted from the initial reference image are utilized as a global anchor and are fed into the core network structure. Furthermore, we integrate a Temporal Motion Block, as introduced in \cite{guo2023animatediff}, to guarantee seamless frame-to-frame transitions and overall video smoothness.

Our network's fundamental architecture is rooted in Latent Diffusion Models (LDMs) \cite{rombach2022high}, integrating a Variational Autoencoder (VAE) \cite{van2017neural} for encoding and decoding, along with a U-Net \cite{ronneberger2015u} architecture. The VAE component is instrumental in mapping input images into a compact latent space, thereby streamlining computations. The U-Net, on the other hand, accepts noise and conditional inputs to forecast the output noise, facilitated by employing a Clip image encoder \cite{radford2021learning} that transforms the reference image into high-dimensional features. These features are then utilized within the U-Net for cross-attention calculations. During training, we intentionally inject noise into the image frames, tasking the network with predicting this added noise. The optimization goal is formally established as follows:

\begin{equation}
  \mathcal{L} = \mathbb{E}_{\mathbf{z}_t,c,\epsilon,t}(\left\|\epsilon-\epsilon_\theta(\mathbf{z}_t,c,t)\right\|),
\end{equation}
where $\mathbf{z}_t$ is the input noise, $c$ represents the condition, $t$ is the timestep. 

Aside from adopting the conventional configurations prevalent in U-Net-based diffusion models, we have devised several bespoke modules to specifically cater to learning both foreground and background dynamics. These modules encompass a Movement Representation system for capturing motion characteristics and a Progressive Long Video Generation mechanism, enabling the creation of extended video sequences.

\begin{figure}[h]
  \centering
  \resizebox{1\linewidth}{!}{
    \includegraphics{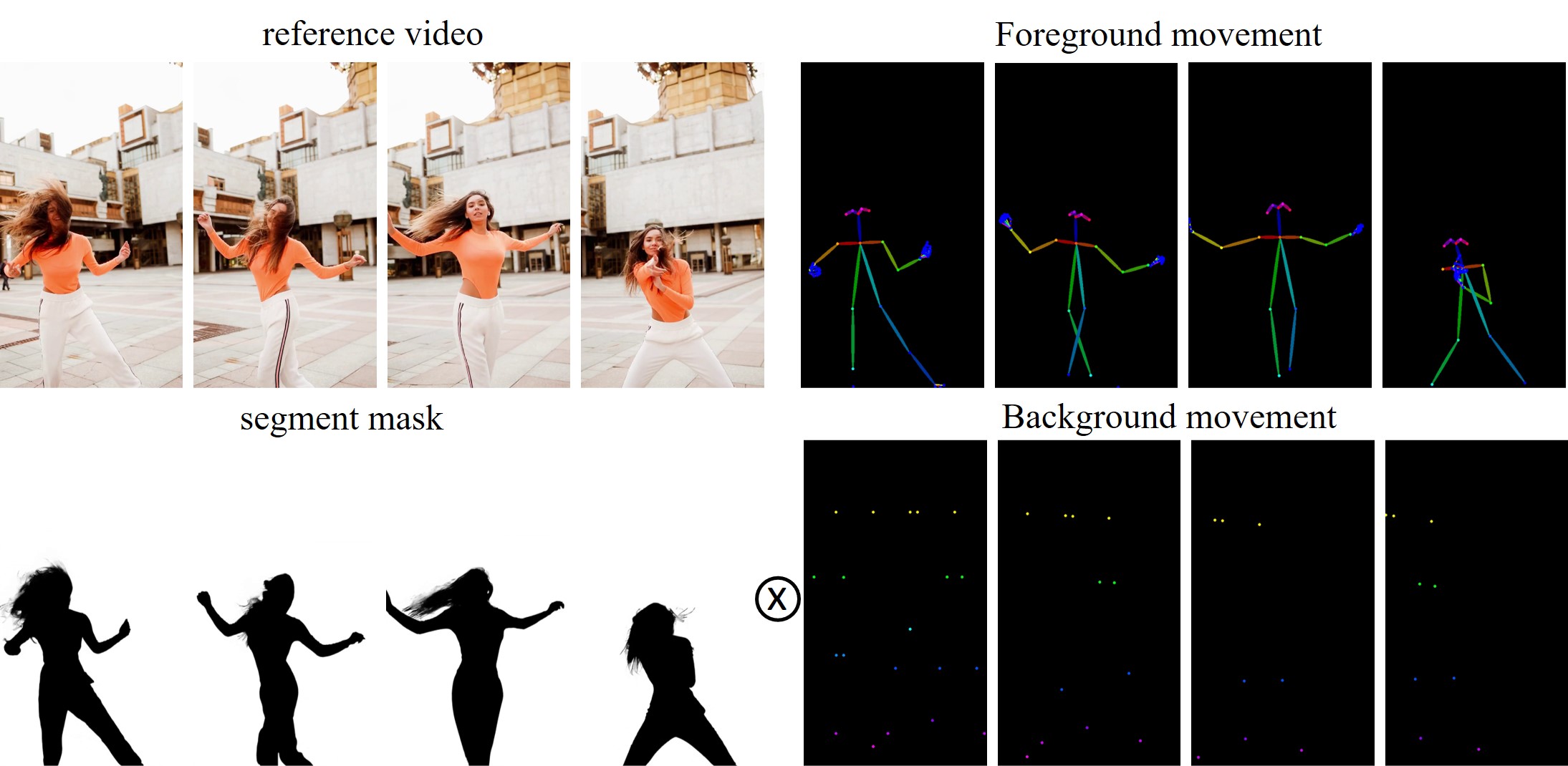} }
  \caption{We separately extract body poses and tracking points from the reference video, with body poses encapsulating the movements of the foreground subjects and tracking points signifying the dynamics of the background. In order to address any potential overlap between these two distinct representations, we employ the inverse of the foreground mask, which is multiplied with the extracted tracking points. This operation effectively removes any overlap, ensuring a clean separation of foreground and background motion elements.}
  \label{fig:demo}
\end{figure}

\subsection{Movement Representation}
Previous research predominantly concentrates on rendering foreground activity through pose-guided frameworks, often disregarding the concurrent background motion. In genuine video content, backgrounds typically shift in harmony with the foreground actions. Foreground human movements can be arbitrary, intertwining bodily gestures and alterations in camera perspective. Meanwhile, the motion of the background can stem from initial camera maneuvers or subsequent video post-production edits, rendering the overall movement intricate. Acknowledging the divergent characteristics between foreground and background motions, we suggest isolating their representations. Estimating camera movement directly from videos where the foreground is dynamically in motion proves imprecise; hence, we introduce a method utilizing tracking points to articulate background motion. These points meticulously record the positions of identical elements across sequential frames, furnishing a precise illustration of the true background dynamics. It is noteworthy that the intricacy of human bodily movements poses considerable difficulties for tracking. In alignment with established literature, we adopt the use of keypoint-based poses to analyze human motion. Specifically, we harness DWPose \cite{yang2023effective}, a tool proven effective in extracting pose details from both training and reference footage, thereby efficiently encapsulating the complexities inherent in human kinetics.

Cotracker \cite{karaev2023cotracker} is employed to extract tracking points from video content. To bolster the robustness of our proposed method against variations in the number of tracking points, we randomly select a subset ranging from 5 to 50 points for each training video. Considering that the extraction of foreground poses and background tracking points is carried out through separate methods, instances may arise where these pose keypoints and tracking points intersect. To mitigate such overlaps, we undertake a meticulous segmentation of the foreground, applying a mask to the background representation to eliminate these superimpositions. We adopt a sophisticated human matting algorithm \cite{liu2020boosting} to yield an accurate alpha mask that delineates the contours of the foreground figure with precision. This way, the resultant depiction of background motion is attained by combining the inverted foreground mask with the extracted tracking points, thereby assuring a cohesive and non-interfering portrayal of movement for both foreground and background components.

\begin{equation}
  \label{alpha_blending}
  M = (1- \alpha)\times T,
\end{equation}
where $T$ denotes the tracked key points, and $\alpha$ signifies the computed foreground mask, we illustrate this procedural workflow in Figure \ref{fig:demo}.

We utilize dual encoder networks, specifically a foreground encoder and a background encoder, as illustrated in Figure \ref{fig:pipeline}, to separately derive features from the representations of foreground and background motions. These extracted motion features—pertaining to both the foreground and background—are then added to features obtained from an input noise vector and a conditional image. The consolidated set of features is subsequently fed into the primary U-Net architecture, facilitating comprehensive motion synthesis that accounts for both contextual and stochastic elements.

\subsection{Progressive Long Video Generation}
 Constrained by the finite capacity of GPU memory, our model's training is confined to a restricted number of frames. During singular inference instances, our methodology facilitates the generation of up to 32 consecutive frames. To extend this capability and produce lengthier videos, approaches like those outlined in \cite{xu2023magicanimate} and \cite{hu2023animate} have adopted a sliding window technique, allowing clips to overlap. This strategy proves efficacious when dealing with static background; however, it encounters substantial limitations when applied to scenarios involving dynamic background motion. In the presence of background movement, the disparity between successive clips becomes more pronounced compared to scenes with a stationary background. Consequently, averaging operations conducted in the latent space under these conditions can induce blurriness and inconsistency in the generated video sequences, undermining visual coherence and quality.
 
Addressing this challenge, we introduce a novel strategy aimed at producing coherent outputs directly between successive video segments. Initiation of the video generation begins with the initial reference image serving as the conditional input for the first clip. Progressing onwards, we adopt a mechanism where the terminal frame of each preceding clip acts as the conditioning input for the succeeding one, as illustrated in Figure \ref{fig:pipeline}. To ascertain the alignment of the generated sequence, a condition concatenation technique is employed, thereby positioning the conditional input image as the opening frame of the output sequence. Consequently, the initial frame of any subsequent clip seamlessly aligns with the finale of its antecedent, fostering continuity. Anticipating potential accumulative errors throughout this iterative procedure, we devise a solution entailing the extraction of global features from the initial reference image. These features are then consistently infused throughout the entire generative process. This measure bolsters the preservation of visual consistency, empowering our model to synthesize videos extending beyond 10 seconds without compromising on coherence or quality.

 \paragraph{Condition concatenation.} 
 
 In prior studies \cite{hu2023animate,karras2023dreampose,xu2023magicanimate,wang2023disco}, the produced frames are primarily governed by the input pose directives and might not closely resemble the conditioning image in detail. We introduce an approach where the image latent representation of the conditional image is directly concatenated with the noise latent, forming the input to the U-Net architecture. By implementing this methodology, the network is encouraged to initiate frame generation with an explicit directive, ensuring that the output sequence begins precisely with the conditioned image, thereby enhancing visual coherence and control over the synthesis process. For the situation where the initial pose guide frame doesn't match the reference image, we substitute a blank image for generating the first video clip. When generating consecutive clips, we employ the last frame of previous clip for condition concatenation.

\paragraph{Global feature extraction and injection.} Within the progressive generation sequence, we adopt the strategy of using the most recently generated frame as the basis for subsequent iterations. However, as observed by \cite{oh2023mtvg}, this practice can lead to error accumulation, causing marked deviations in image content over time. To tackle this issue, we propose incorporating features derived from the initial reference image as global features, infusing them throughout the entire generation workflow. These persistent global features act as a unifying force, maintaining content and stylistic consistency across different stages of generation and effectively mitigating error build-up. To ensure the capture of rich detail from the input image, a potent feature extractor is imperative. While a clip image encoder is capable of extracting high-level features, these may lack the finer nuances necessary for precision. Although Anydoor \cite{chen2023anydoor} employs DINOV2 \cite{oquab2023dinov2} to retrieve more detailed features, these still fall short in high-frequency detail representation. Notably, U-Net architectures have demonstrated exceptional capability in extracting fine, detailed features, as evidenced by various studies including \cite{hedlin2024unsupervised,luo2024diffusion,hu2023animate,xu2024ootdiffusion,zhu2024champ,wei2024aniportrait,tian2024emo,chen2024magic}. To harness this strength, we employ an identical U-Net for the extraction of global features, subsequently integrating these features into the primary network infrastructure using concatenation of self-attention features. This strategic utilization of U-Net not only ensures a rich feature extraction process but also reinforces the consistency and fidelity of the generated content over prolonged sequences. 

\section{Experiments}
\subsection{Implementation Details}
\paragraph{Network training.} The U-Net's weights, along with the global feature extraction component, are initialized leveraging a pretrained stable diffusion model \cite{runwayml2023sd}. All network modules except for VAE encode and decoder, undergo joint training, adopting a learning rate of $1e^{-5}$. To facilitate this training process, we employ a setup consisting of 4 NVIDIA A100 GPUs, with a batch size set to 4. The chosen resolution for training is $640\times 448$ pixels, and to economize on memory usage, we implement gradient checkpointing techniques.

\paragraph{Dataset.} Current publicly available human motion datasets include the TikTok dataset \cite{jafarian2021learning} and the UBC Fashion dataset \cite{zablotskaia2019dwnet}, both of which consist of videos meticulously handpicked for their content. Notably, every video within these datasets features a static background across both training and testing subsets. Conversely, videos captured in real-world scenarios using handheld devices like smartphones characteristically incorporate dynamic backgrounds. In recognition of this discrepancy, we curated the Human-5000 dataset by gathering unaudited, genuine videos from the web, thereby encompassing a diverse array of background movements. This dataset, focusing on human-centric videos, was sourced ethically from copyright-free platforms. Unlike the aforementioned datasets, the Human-5000 does not impose manual curation on the foreground or background motion, allowing for a broad spectrum of naturalistic movements in both elements. The dataset is systematically divided into two parts: a training set comprising 4000 videos and a test set consisting of 1000 videos, with this allocation performed randomly to ensure unbiased evaluation of our model's performance across varied motion scenarios.

\paragraph{Baseline.} Our primary comparative analysis focuses on similar methodologies, notably Disco \cite{wang2023disco}, Magicanimate \cite{chen2024magic}, and Animateanyone \cite{hu2023animate}. We undertake a comprehensive evaluation of these approaches, encompassing both qualitative assessments that gauge visual quality and quantitative measures for objective comparison. Considering that all the mentioned methods have referenced the TikTok dataset \cite{jafarian2021learning}, we initially pit our method against theirs on this common ground, providing a consistent benchmark for comparison. Furthermore, given the unique challenges posed by dynamic backgrounds, we extend our evaluation by contrasting our approach with the state-of-the-art Animateanyone \cite{hu2023animate} on our newly compiled Human-5000 dataset, which specifically incorporates a wide range of background movements. This additional comparison highlights the efficacy of our method in handling complex, real-world scenarios beyond static backgrounds.

\begin{figure}[h!]
  \centering
  \resizebox{1\linewidth}{!}{
    \includegraphics{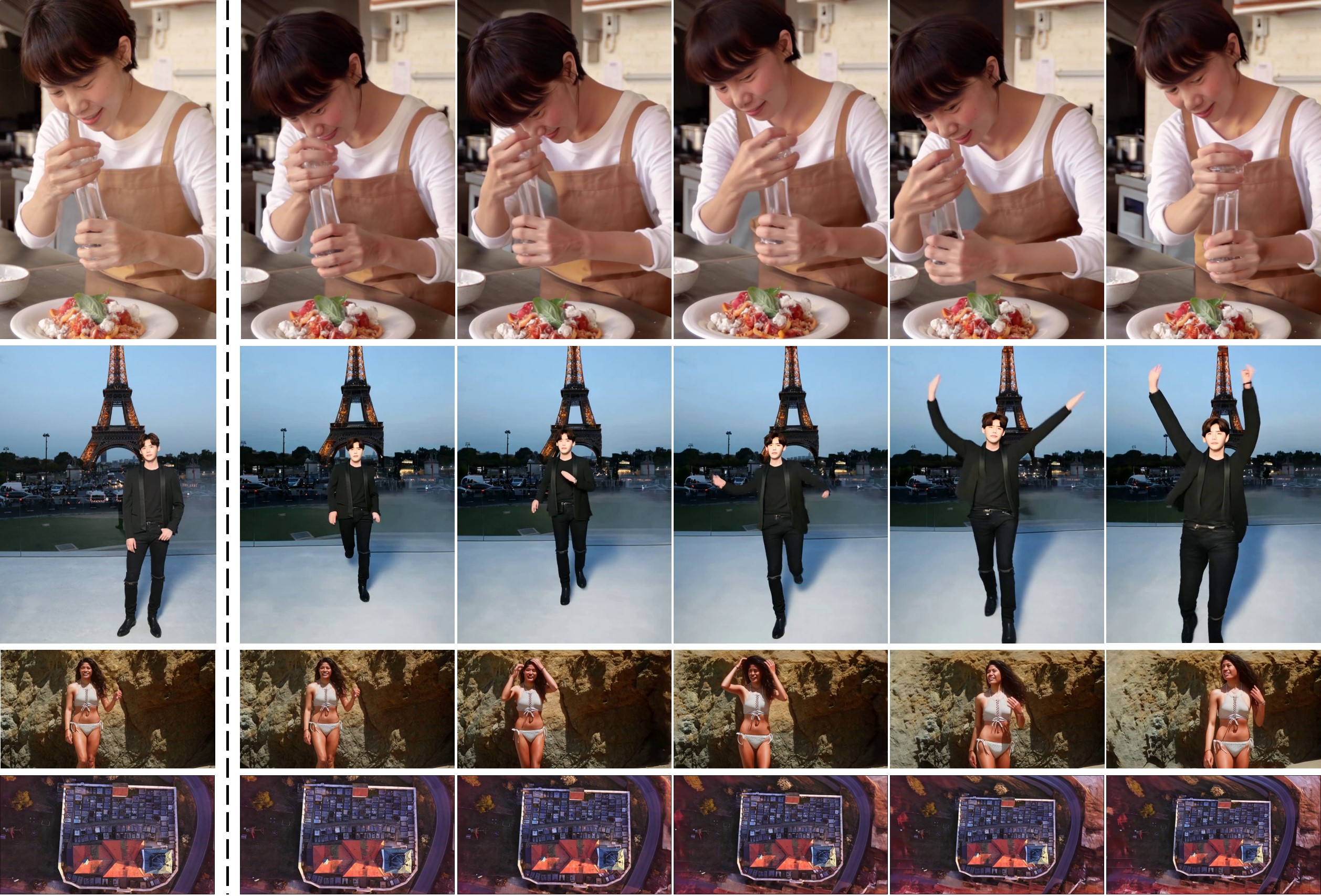} }
  \caption{Inputting previously unseen data, our method's qualitative performance highlights its success in generating videos that seamlessly integrate foreground and background motion.}
  \label{fig:result1}
\end{figure}

\subsection{Results and Comparisons}

In Figure \ref{fig:result1}, We demonstrate qualitative examples of outputs generated by our proposed method using unseen data not part of either the TikTok or Human-5000 datasets. Our technique effectively produces lifelike human videos, marked by natural foreground motion and believable background actions.

\begin{figure}[h!]
  \centering
  \resizebox{1\linewidth}{!}{
    \includegraphics{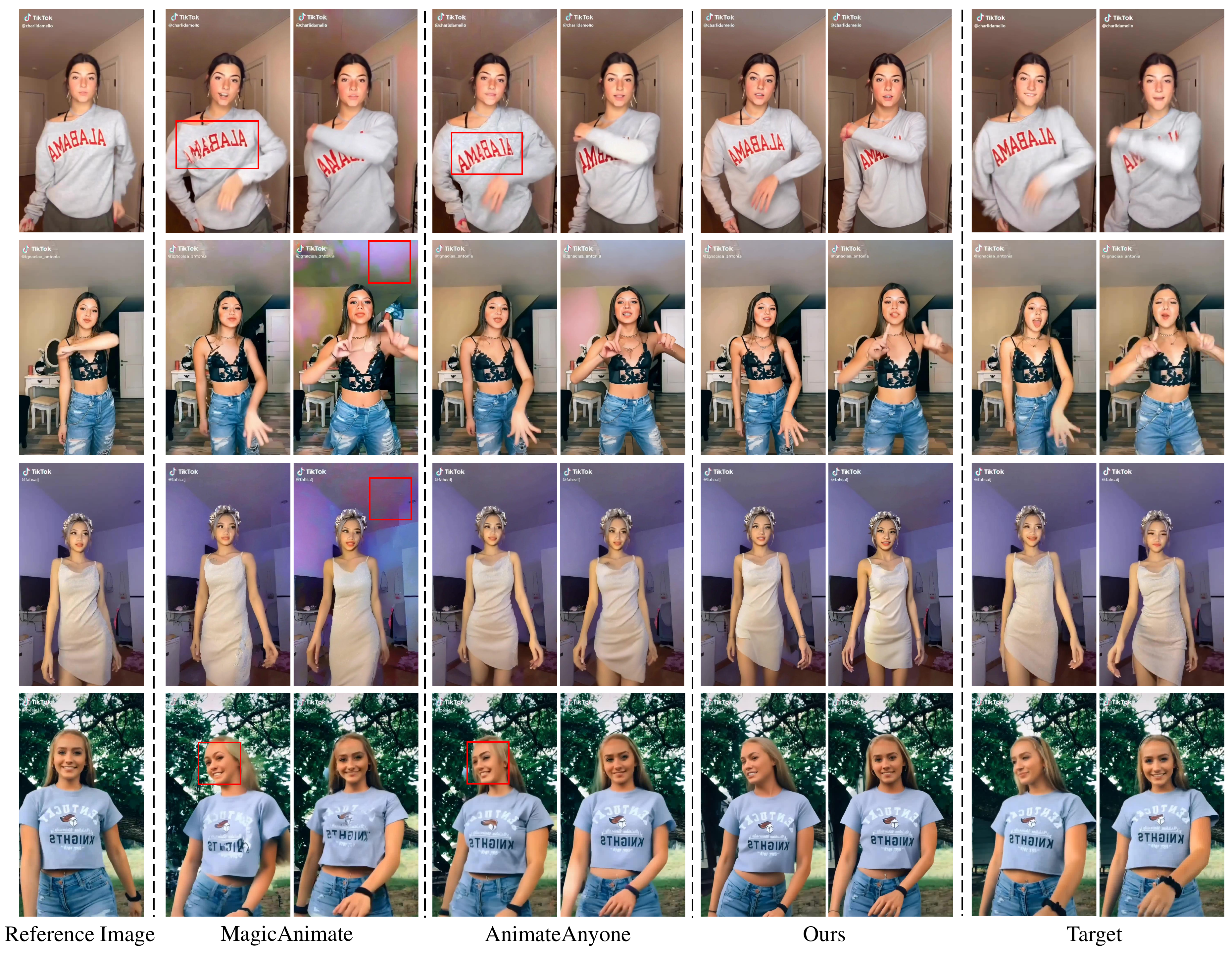} }
  \caption{On the TikTok dataset, qualitative analysis reveals that MagicAnimate \cite{xu2023magicanimate} outputs suffer from color distortion, while AnimateAnyone's \cite{hu2023animate} also show detail inconsistencies like face and text. Our method notably enhances these aspects.}
  \label{fig:tiktok_test}
\end{figure}

We employ the TikTok dataset \cite{jafarian2021learning}, comprising 440 individual dance videos each lasting 10 to 15 seconds. Out of these, 50 videos constitute our test set, and the rest form the training set. All comparative algorithms are trained on the identical training subset. DisCo \cite{wang2023disco} and MagicAnimate \cite{chen2024magic} leverage the pre-trained models provided by their respective authors for initialization, whereas AnimateAnyone \cite{hu2023animate} and our model utilize a pre-trained stable diffusion model for this purpose. A quantitative assessment summarized in Table \ref{tiktok_qc} was conducted, and qualitative comparisons with MagicAnimate \cite{chen2024magic} and AnimateAnyone \cite{hu2023animate} are illustrated in Figure \ref{fig:tiktok_test}. Both the qualitative and quantitative evaluations confirm the efficacy of our approach, with our method outperforming prior works in frame quality and level of detail.

\begin{table}
  \caption{Quantitative comparison on TikTok dataset}
  \label{tiktok_qc}
  \centering
  \begin{tabular}{lllll}
    \toprule
     Methods  & SSIM$\uparrow$ & PSNR$\uparrow$ & LPIPS$\downarrow$ & FVD$\downarrow$ \\
    \midrule
    DisCo \cite{wang2023disco} &  0.665 &    28.60 &   0.299  &     302.2  \\
    MagicAnimate \cite{xu2023magicanimate} &  0.714 &   29.16  &   0.253  &   182.1        \\
    AnimateAnyone \cite{hu2023animate} &  0.718 &   \textbf{29.56}  &   0.278  &    171.9   \\
    Ours     &  \textbf{0.738} &    29.43 &    \textbf{0.227} &     \textbf{168.2} \\
    \bottomrule
  \end{tabular}
\end{table}

\begin{figure}[h!]
  \centering
  \resizebox{1\linewidth}{!}{
    \includegraphics{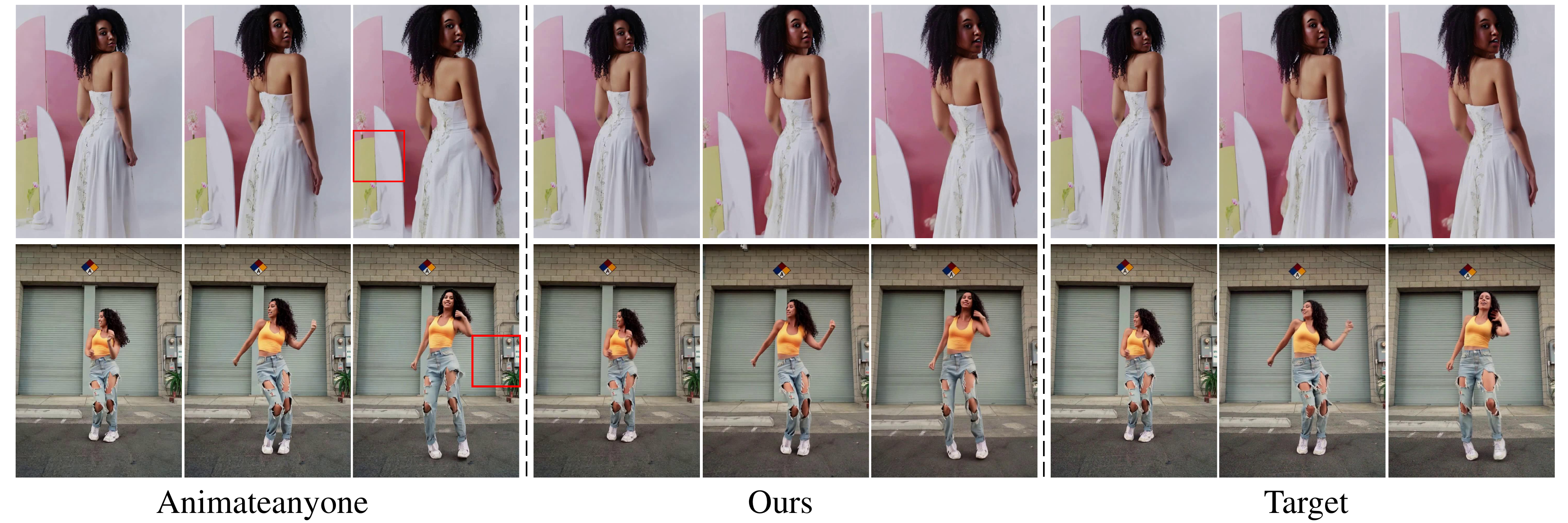} }
  \caption{When compared with AnimateAnyone (\cite{hu2023animate}) in Human-5000 datasets featuring background motion, our method excels with superior background movement generation capabilities.}
 \label{fig:compare_aa}
\end{figure}

In Figure \ref{fig:compare_aa} and Table \ref{Human-5000_qc}, we specifically pit our method against Animateanyone \cite{hu2023animate}, utilizing our custom Human-5000 dataset as the testing arena. This unique dataset is distinguished by its inclusion of videos characterized by concurrent foreground and background motion. We observe that Animateanyone struggles to produce convincing videos when confronted with dynamic backgrounds. Conversely, our methodology excels in generating videos that seamlessly integrate natural and fluid background movements, highlighting its superiority in managing complex, real-world motion scenarios.

\begin{table}
  \caption{Quantitative comparison on Human-5000 dataset}
  \label{Human-5000_qc}
  \centering
  \begin{tabular}{lllll}
    \toprule
     Methods  & SSIM$\uparrow$ & PSNR$\uparrow$ & LPIPS$\downarrow$ & FVD$\downarrow$ \\
    \midrule
    AnimateAnyone\cite{hu2023animate} &  0.839 &    29.31 &   0.159  &     201.2  \\
    \midrule
    W/O foreground representation &  0.828 &    25.81 &   0.216  &     464.9 \\
    W/O background representation &  0.842 &    29.11 &   0.160  &     197.8 \\
    W/O gloabl feature &  0.847&    27.87 &   0.184  &     313.0 \\
    W/O condition concatenation &  0.829 &    27.96 &   0.168  &     204.1 \\
    \midrule
    Ours     &  \textbf{0.866} &    \textbf{29.69} &    \textbf{0.139} &     \textbf{182.6} \\
    \bottomrule
  \end{tabular}
\end{table}

\begin{figure}[h!]
  \centering
  \resizebox{1\linewidth}{!}{
    \includegraphics{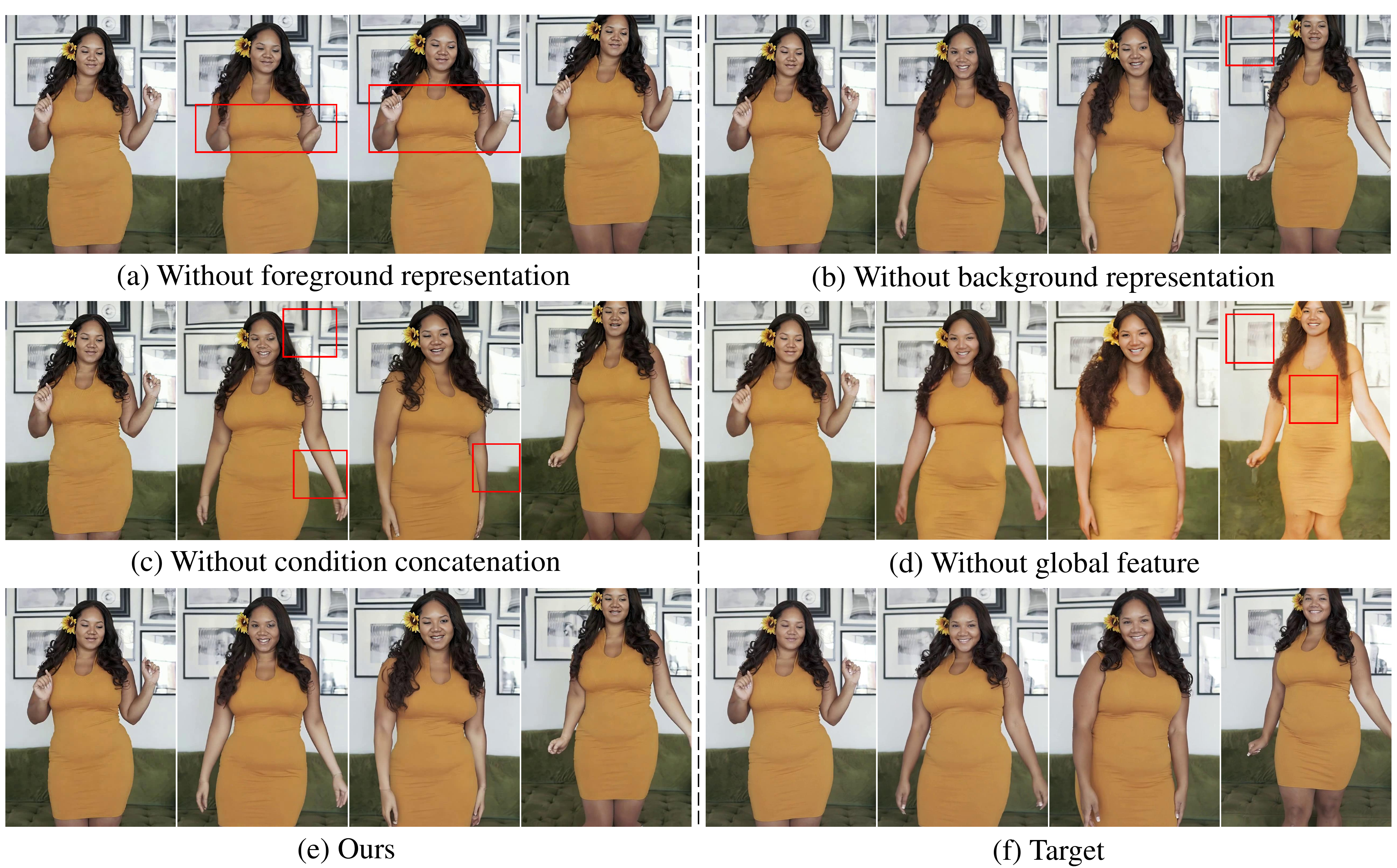} }
  \caption{Ablation study with different settings.}
 \label{fig:ablation}
\end{figure}

\subsection{Ablation Study}

\paragraph{Without foreground movement representation.} Pose information is incorporated to better illustrate the movement of foreground humans since tracking points alone prove insufficient in accurately capturing the intricacies of bodily motion. As a point of comparison, we conduct experiments omitting foreground representation and relying solely on tracking points to convey motion. The resultant data, depicted in Figure \ref{fig:ablation}(a) and quantified in Table \ref{Human-5000_qc}, reveal a deficiency in effectively generating realistic human body movements.

\paragraph{Without background movement representation.} We introduce the use of sparse tracking points as a means to encode background motion during both training and inference phases. To validate the efficacy of this component, we conduct an ablation study by re-training our network sans the background representation module. As evidenced in Figure \ref{fig:ablation}(b) and Table \ref{Human-5000_qc}, the resultant videos produced by this modified network lack any discernible background movement, leading to outputs that appear less authentic and lifelike. This comparison underscores the pivotal role of background representation in enhancing the realism of our generated videos.

\paragraph{Without condition concatenation.} Condition concatenation serves as a bridge, seamlessly linking the terminal frame of one video clip to the initial frame of the succeeding one. To illustrate its necessity, Figure \ref{fig:ablation}(c) presents examples generated without employing condition concatenation. For the generation of extended sequences, we adopt the sliding window technique, as utilized in \cite{xu2023magicanimate} and \cite{hu2023animate}. Evidently, the absence of condition concatenation impedes the attainment of smooth frame transitions between clips, underscoring its importance in ensuring contiguous and fluid video sequences.

\paragraph{Without global feature extraction.} The integration of a global feature extraction component is pivotal in maintaining content and style uniformity throughout the progression of lengthy video generation. To assess its impact, we undertake an experiment where we omit this component and retrain the network using identical training configurations. The outcomes depicted in Figure \ref{fig:ablation}(d) and Table \ref{Human-5000_qc} illustrate that, in the absence of global feature extraction, the video content exhibits marked inconsistencies and notable alterations across consecutive clips. Conversely, when the global feature extraction is included, the resultant video demonstrates remarkable consistency and retains a high level of visual quality, emphasizing its critical contribution to coherent, high-fidelity video synthesis.

\section{Limitations and Discussions}
In our movement representation approach, we employ DWPose \cite{yang2023effective} for extracting human poses and Cotracker \cite{karaev2023cotracker} to identify background motion. Nonetheless, our method's performance is contingent upon the accuracy of these extractions; inaccuracies or omissions in the extracted movements by either technique could undermine the final output.

Furthermore, given the sparsity of tracking points allocated for background motion, there exists the possibility that the full breadth of background dynamics may not be comprehensively captured by these points alone. Consequently, the synthesized video might only reflect a portion of the background movement present in the reference video. This shortcoming can be somewhat alleviated by increasing the number of tracking points dedicated to background motion representation, thereby enhancing the system's capacity to replicate intricate background activities more faithfully.

Our method excels in producing lifelike videos. However, it's noteworthy that advancements in generative model capabilities could pave the way for deepfake production, posing risks of misuse for disseminating false information. On the flip side, the absence of audio in our generated videos serves as a distinguishing feature, facilitating their differentiation from genuine footage.

\section{Conclusion}
This paper introduces a novel approach to video generation that segregates the representation of foreground and background motion. By adopting pose estimation for foreground dynamics and sparse tracking points for background movement, our method achieves the creation of videos that exhibit both natural human action and authentic environmental motion, surpassing the constraints of prior techniques confined to static backgrounds. Moreover, we present an innovative pipeline designed to facilitate the synthesis of extended video sequences without encountering cumulative errors over time. Central to this advancement are techniques of condition concatenation and global feature extraction, which collectively empower the generation of prolonged, clip-by-clip video content while preserving visual coherence and consistency throughout.

\bibliographystyle{abbrv}
\bibliography{ref}

\end{document}